\documentclass[accepted]{car2025} % for initial submission
\usepackage[american]{babel}
\usepackage{natbib} % has a nice set of citation styles and commands
\usepackage{mathtools} % amsmath with fixes and additions
\usepackage{booktabs} % commands to create good-looking tables
\usepackage[T1]{fontenc}    % use 8-bit T1 fonts
\usepackage{hyperref}       % hyperlinks
\usepackage{url}            % simple URL typesetting
\usepackage{booktabs}       % professional-quality tables
\usepackage{amsfonts}       % blackboard math symbols
\usepackage{nicefrac}       % compact symbols for 1/2, etc.
\usepackage{microtype}      % microtypography
\usepackage{xcolor}         % colors
\usepackage{mathtools}
\usepackage{amsmath,amsfonts,amssymb,dsfont,bm,amsthm}
\usepackage{csquotes}
\usepackage{float}
\usepackage{subcaption}
\usepackage{txfonts}
\usepackage{dsfont}
\usepackage{hyperref}
\usepackage{thm-restate}
\usepackage{soul}
\usepackage{fix-cm}
\usepackage{comment}
\usepackage{soul}
\usepackage{tikz} % nice language for creating drawings and diagrams\usepackage[utf8]{inputenc} % allow utf-8 input
\usetikzlibrary{babel, arrows, decorations, decorations.pathmorphing, decorations.pathreplacing, decorations.shapes, calc,arrows.meta, positioning, decorations.markings,fit,shapes.geometric}

\newtheorem{theorem}{Theorem}
\newtheorem{lemma}{Lemma}
\newtheorem{corollary}{Corollary}

\newtheorem{definition}{Definition}
\newtheorem{example}{Example}

\usepackage{slashed}
\newcommand{\ind}{{\perp\!\!\!\perp}}
\newcommand{\leftcomingarrow}{{\leftarrow \! \! \! \! \bullet}}
\newcommand{\rightcomingarrow}{\bullet \! \! \! \! \rightarrow}
\newcommand{\notind}{\slashed{\ind}}
\newcommand{\V}{\mathcal{V}}
\newcommand{\F}{\mathcal{F}}
\newcommand{\E}{\mathcal{E}}
\newcommand{\X}{\mathbf{X}}
\newcommand{\Y}{\mathbf{Y}}
\newcommand{\Z}{\mathbf{Z}}
\newcommand{\W}{\mathbf{W}}
\newcommand{\Pa}{\text{Pa}}
\newcommand{\Anc}{\text{Anc}}
\newcommand{\Desc}{\text{Desc}}
\newcommand{\G}{{\mathcal{G}}}
\newcommand{\M}{{\mathcal{M}}}
\newcommand{\C}{\mathcal{C}}
\newcommand{\U}{\mathcal{U}}
\newcommand{\Do}{\text{do}}

\newcommand{\Ch}{\text{Ch}}

\title{Identifiability in Causal Abstractions: A Hierarchy of Criteria}

\author[1]{Clément Yvernes}
\author[1]{Emilie Devijver}
\author[2]{Marianne Clausel}
\author[1]{Eric Gaussier}

\affil[1]{%
Univ Grenoble Alpes, CNRS, Grenoble INP, LIG
}
\affil[2]{%
 Université de Lorraine, CNRS, IECL 54000 Nancy
}

\begin{document}
\maketitle

\begin{abstract}
Identifying the effect of a treatment from observational data typically requires assuming a fully specified causal diagram. However, such diagrams are rarely known in practice, especially in complex or high-dimensional settings. To overcome this limitation, recent works have explored the use of causal abstractions—simplified representations that retain partial causal information. In this paper, we consider causal abstractions formalized as collections of causal diagrams, and focus on the identifiability of causal queries within such collections. We introduce and formalize several identifiability criteria under this setting. Our main contribution is to organize these criteria into a structured hierarchy, highlighting their relationships. This hierarchical view enables a clearer understanding of what can be identified under varying levels of causal knowledge. We illustrate our framework through examples from the literature and provide tools to reason about identifiability when full causal knowledge is unavailable.
\end{abstract}

\section{Introduction}\label{sec:intro}

Knowing the effect of a treatment $X$ on an outcome $Y$, encoded by an intervention $do(X)$ in the interventional distribution $P(Y|do(X))$, is crucial in many applications. However, performing interventions is often impractical due to ethical concerns, potential harm or prohibitive costs. In such cases, one can instead aim to identify do-free formulas that estimate the effects of interventions using only   observational (non-experimental) data and  a causal graph \citep{pearl_causality_2009}. Solving the identifiability problem typically involves establishing graphical criteria on causal diagrams under which the total effect is identifiable and providing a do-free formula for estimating it from observational data. However, specifying a causal diagram requires prior knowledge of the  relationships between all observed variables, a requirement that is often unmet in real-world applications. This challenge is particularly acute in complex, high-dimensional settings, limiting the practical applicability of causal inference methods.

A more relaxed assumption than having access to a fully specified causal graph is to rely on a causal abstraction—a representation that captures some, but not necessarily all, of the underlying causal structure. Causal abstraction has been recently studied from multiple angles: its compositional properties \citep{pmlr-v161-rischel21a}, the search for optimal abstractions \citep{zennaro2022computingoptimalabstractionstructural}, the use of macro-variables or variable groupings in place of micro-level variables \citep{PARVIAINEN2017110, DBLP:journals/corr/abs-1812-03789}, and the quantification of information loss relative to the full causal model \citep{10.24963/ijcai.2023/638}. Abstractions have also been applied across a variety of contexts, including causal discovery \citep{WahlNinadRunge+2024}, query identification \citep{anand_causal_2023}, reinforcement learning \citep{10.5555/3702676.3702869}, as well as structured data settings such as time series \citep{assaad2022, 10.5555/3702676.3702683, multivariateICA}, neural networks \citep{NEURIPS2021_4f5c422f}, mechanistic interpretability \citep{geiger2025causalabstractiontheoreticalfoundation} and hierarchical data \citep{weinstein2024hierarchicalcausalmodels}.

In this work, we formalize a causal abstraction as a collection of causal diagrams, and focus on the problem of identifying a causal query within such a collection. While various notions of identifiability have been proposed in this context, there is currently no unified framework that connects them. Our goal is to clarify and relate these different notions by introducing a formal structure for reasoning about identifiability in abstracted causal settings. More precisely, our contributions are as follows:
\begin{itemize}
\item We formally define several notions of identifiability directly on collections of graphs;
\item We analyze the relationships between these different notions;
\item We illustrate each concept with examples and connections to existing work from the literature.
\end{itemize}

The remainder of the paper is structured as follows: Section \ref{sec:preliminaries} introduces the main concepts while Section \ref{sec:singleGraph} presents identifiability in a single graph; and Section \ref{sec:collectionGraphs} presents different notions of identifiability in a collection of graphs. Section \ref{sec:prop} provides the link between those notions of identifiability, and Section \ref{section:conjecture} develops a conjecture. Lastly, Section~\ref{sec:conclusion} concludes the paper.

\section{Preliminaries}
\label{sec:preliminaries}
%In this section, we remind the main concepts and notations.
\paragraph{Graphs.} 
A \emph{(directed) mixed graph} $\G = (\mathcal{V}, \mathcal{E})$ is a graph that contains two type of arrows, \emph{directed arrows} ($\rightarrow$) and \emph{dashed-bidirected arrows} ($\dashleftrightarrow$). Following the notations of \cite{perkovic_complete_2018}, an arrow ($\leftcomingarrow$) represents either a directed arrow ($\leftarrow$) or a dashed-bidirected arrows ($\dashleftrightarrow$). We potentially distinguish arrows $\mathcal{E}$ into directed arrows $\E_D$ and dashed bidirected arrows $\E_B$.
For a mixed graph $\G$, if $X\rightarrow Y$, then $X$ is a \emph{parent} of $Y$ and $Y$ is a \emph{child} of $X$. A \emph{path} is a sequence of distinct vertices such that there exists an arrow in $\G$ connecting each element to its successor. A \emph{directed path}, or a \textit{causal path}, is a path in which all arrows are pointing towards the last vertex.
If there is a directed path from $X$ to $Y$, then $X$ is an \emph{ancestor} of $Y$, and $Y$ is a \emph{descendant} of $X$.\footnote{By convention, a vertex is a descendant and an ancestor of itself.} The sets of parents, children, ancestors and descendants of $X$ in $\G$ are denoted by $\text{Pa}(X,\G)$, $\Ch(X,\G)$, $\Anc(X,\G)$ and $\Desc(X,\G)$ respectively. 
For two disjoint subsets \(\X, \Y \subseteq \V\), a \textit{path} from $\X$ to $\Y$ is a path from some \(X \in \X\) to some  \(Y \in \Y\). A path from $\X$ to $\Y$ is \textit{proper} if only its first node is in $\X$.
A \emph{directed cycle} is a circular list of distinct vertices in which each vertex is a parent of its successor. When a (directed) mixed graph does not contain any directed cycle, then this graph is called an \emph{Acyclic Directed Mixed Graph (ADMG).} When an ADMG does not contain any dashed bidirected arrow, it is called a \emph{Directed Acyclic Graph (DAG)}. A graph \( \G_1 = (\V_1, \E_1) \) is said to be a subgraph of another graph \( \G_2 = (\V_2, \E_2) \) if \( \V_1 \subseteq \V_2 \) and \( \E_1 \subseteq \E_2 \). In this case, we write \( \G_1 \subseteq \G_2 \), which defines a partial order over graphs.

On a path $\pi$, a vertex $V$ is said to be a \emph{collider} on $\pi$ if it has two incoming arrows (i.e. $\rightcomingarrow V \leftcomingarrow$) in $\pi$. Otherwise, it is said to be a \emph{non-collider} on $\pi$.
A vertex $V$ is said to be \emph{active} on a path relative to $\Z$ if 1) $V$ is a collider and $V$ or any of its descendants are in $\Z$ or 2) $V$ is a non-collider and is not in $\Z$. A path $\pi$  is said to be \emph{active} given (or conditioned on) $\Z$ if every vertex on $\pi$ is active relative to $\Z$. Otherwise, $\pi$   is said to be \emph{inactive} given $\Z$. Given a graph $\mathcal{G}$, the sets $\X$ and $\Y$ are said to be \emph{d-separated} by $\Z$ if every path between $\X$ and $\Y$ is inactive given $\Z$. We denote this by $\X \ind_{\G} \Y \mid \Z$. Otherwise, $\X$ and $\Y$ are said to be \emph{d-connected} given $\Z$, which we denote by $\X \notind_{\G} \Y \mid \Z$.

The mutilated graph ${\mathcal{G}}_{\overline{\X}\underline{\Z}}$ is the result of removing from a graph ${\mathcal{G}}$ edges 
with an arrowhead into $\X$ (e.g., $A \rightcomingarrow \X$),  
and edges with a tail from $\Z$ (e.g., $A \leftarrow \Z$).

\paragraph{Structural Causal Models.} A \emph{{Structural Causal Model} (SCM)} $\mathcal{M}$ is a 4-tuple $\langle \U, \V, \F_\M, P(\U)\rangle$, where $\U$ is a set of exogenous (latent) mutually independent variables and $\V$ is a set of endogenous (measured) variables. $\F_\M$ is a collection of functions $\{f_i\}_{i=1}^{|\V|}$ such that each endogenous variable $V_i\in\V$ is a function $f_i\in\F_\M$ of $\U_i\cup \Pa(V_i)$, % to $V_i$,
where $\U_i\subseteq\U$ and $\Pa(V_i)\subseteq\V\setminus V_i$. The uncertainty is encoded through a probability distribution over the exogenous variables, $P(\U)$. The resulting joint distribution over \(\V\), obtained by marginalizing out \(\U\), is called the \emph{observational distribution} and is denoted \(P_{\mathcal{M}}\left(\V\right)\).
Each SCM $\mathcal{M}$ induces an ADMG -- $\G_{\mathcal{M}} \coloneqq (\V, \E = (\E_{D},\E_{B}))$, known as a \emph{causal diagram}, that encodes the structural relations among $\V\cup\U$, where every $V_i \in \V$ is a vertex. 
$\G_{\mathcal{M}}$ contains directed edges which connect each variable $V_i \in \mathcal{V}$ to its parents $V_j \in \Pa(V_i)$ as $(V_j \rightarrow V_i)$, and bidirected edges between variables $V_i, V_j \in \mathcal{V}$ that share a common exogenous parent, i.e., such that $\mathcal{U}_i \cap \mathcal{U}_j \neq \emptyset$, denoted $V_i \dashleftrightarrow V_j$.

\paragraph{Intervention.}
In a given SCM \(\mathcal{M}=\langle \U,\V,\mathcal{F},P(\U)\rangle\), let \(\mathbf{V}=(V_1,\dots,V_n) \) be a subset of \(\V\). An intervention \(\Do(\mathbf{V}=\mathbf{v})\) replaces each structural function \(f_i\) for \(V_i\in\mathbf{V}\) by the constant assignment \(V_i:=v_i\), which in the causal diagram \(\G_{\mathcal{M}}\) corresponds to removing all incoming edges into each \(V_i\). The resulting joint distribution over \(\V\), obtained by marginalizing out \(\U\), is called the \emph{interventional distribution} and is denoted \(P_{\mathcal{M}}\left(\V\mid\Do(\V =\mathbf{v})\right)\) or \(P_{\mathcal{M}}\left(\V\mid\Do(\mathbf{v})\right)\) for short. Let $\X$, $\Y$, and $\Z$ be pairwise distinct subsets of $\V$. 

\paragraph{Causal Query and Estimand.} Let $\V$ denote the set of observable variables and let $\X,\Y,\Z\subseteq\V$ be pairwise disjoint. A \emph{causal query} is any expression of the form \( P(\mathbf{y} \mid \Do(\mathbf{x}), \mathbf{z}) \). A \emph{causal estimand} is any function \( \mathcal{F} \) mapping distributions over \( \V \) to \( \mathbb{R}^+ \). Given an SCM \( \mathcal{M} = \langle \U, \V, \mathcal{F}, P(\U) \rangle \), $P_{\M}(\V)$ its observational distribution and a causal query \( P(\mathbf{y} \mid \Do(\mathbf{x}), \mathbf{z}) \), a causal estimand \( \mathcal{F} \) is said to be \emph{valid} in \( \mathcal{M} \) for \( P(\mathbf{y} \mid \Do(\mathbf{x}), \mathbf{z}) \) if it satisfies  
\(
\mathcal{F}\left(P_{\mathcal{M}}(\V)\right) = P\left(\mathbf{y} \mid \Do(\mathbf{x}), \mathbf{z}\right).
\)

\begin{comment}
\paragraph{Causal Bayesian Network.}
Let $\G$ be a causal diagram over variables $\V$, and let $P(\V)$ be a joint distribution on $\V$.  Define
\(
P_\star \;=\;\left\{P(\V\mid \Do(\mathbf{x}))\right\}_{\X\subseteq\V,\;\mathbf{x}}
\)
as the family of all interventional distributions.  We say that $\G$ is a \emph{causal Bayesian network} compatible with $P_\star$ if, for every subset $\mathbf{X}\subseteq\V$, every intervention value $\mathbf{x}$, and every full assignment $v$ of $\V$, 
\[
P\left(v\mid \Do(\mathbf{x})\right)
=
\begin{cases}
    \displaystyle
    \prod_{i\,:\,V_i\notin\mathbf{X}}P\left(v_i\mid \mathrm{pa}(v_i)\right)
    &\text{if $v$ agrees with $\mathbf{x}$,}\\[0.5ex]
    0 &\text{otherwise,}
\end{cases}
\]
where $\mathrm{pa}(v_i)$ denotes the values assigned to the parents $\Pa(V_i)$ in $v$.  In particular, any causal diagram and interventional distributions arising from an SCM satisfy this condition by construction.
\end{comment}

\section{Identifiability in a Single Graph}
\label{sec:singleGraph} 
Different structural causal models (SCMs) can induce the same causal diagram. Consequently, a causal diagram represents a collection of SCMs. A causal query 
\(
  P\bigl(\mathbf{y} \mid \Do(\mathbf{x}), \mathbf{z}\bigr)
\)
is said to be \emph{identifiable} if it can be uniquely determined from the observational distribution.

\begin{definition}[Identifiability in a Single Graph]\label{def:single}
    Let $\G = (\V,\E)$ be a causal diagram, and let $\X,\Y,\Z\subseteq \V$ be pairwise disjoint. The causal query 
    \(P\bigl(\mathbf{y} \mid \Do(\mathbf{x}),\mathbf{z}\bigr)\)
    is \emph{identifiable} in $\G$ if there exists a causal estimand valid for \(P\bigl(\mathbf{y} \mid \Do(\mathbf{x}),\mathbf{z}\bigr)\) in every  SCM $\M$ satisfying \(\G_{\M} = \G\). In this case, we also say that the causal estimand is valid for \(P\bigl(\mathbf{y} \mid \Do(\mathbf{x}),\mathbf{z}\bigr)\) in $\G$.
\end{definition}

Pearl’s do-calculus provides a set of three rules for transforming interventional distributions into observational ones. We restate these rules below.

\begin{theorem}[Do-Calculus {\citep{pearl_causality_2009}}]
\label{th:do_calculus}
Let $\G=(\V,\E)$ be a causal diagram, and let $\X,\Y,\Z,\W\subseteq\V$ be pairwise disjoint. For any SCM $\M$ inducing $\G$, denote its interventional density by $P_\M$. Then:

\begin{enumerate}
  \item[R1.] Insertion/deletion of observations. 
    \[
      P_\M\bigl(\mathbf{y} \mid \Do(\mathbf{w}),\,\mathbf{x},\,\mathbf{z}\bigr)
      =
      P_\M\bigl(\mathbf{y} \mid \Do(\mathbf{w}),\,\mathbf{z} \bigr)
    \]
    if 
    \(\Y \;\ind_{\G_{\overline{\W}}}\;\X \mid (\W \cup \Z)\).

  \item[R2.] Action/observation exchange.  
    \[
      P_\M\bigl(\mathbf{y} \mid \Do(\mathbf{w}),\,\Do(\mathbf{x}),\,\mathbf{z} \bigr)
      =
      P_\M\bigl(\mathbf{y} \mid \Do(\mathbf{w}),\,\mathbf{x},\,\mathbf{z} \bigr)
    \]
    if 
    \(\Y \;\ind_{\G_{\overline{\W},\,\underline{\X}}}\;\X \mid (\W \cup \Z)\).

  \item[R3.] Insertion/deletion of actions. 
    \[
      P_\M\bigl(\mathbf{y} \mid \Do(\mathbf{w}),\,\Do(\mathbf{x}),\, \mathbf{z}\bigr)
      =
      P_\M\bigl(\mathbf{y} \mid \Do(\mathbf{w}),\,\mathbf{z} \bigr)
    \]
    if 
    \(\Y \;\ind_{\G_{\overline{\W},\,\overline{\X(\Z)}}}\;\X \mid (\W \cup \Z)\),  
    where 
    \(\X(\Z) := \X \setminus \Anc_{\G_{\overline{\W}}}(\Z)\).
\end{enumerate}
\end{theorem}

A \emph{proof of do-calculus} is a finite sequence of applications of the do-calculus rules, possibly combined with standard probability manipulations, that rewrites a causal query \( P\bigl(\mathbf{y} \mid \Do(\mathbf{x}), \mathbf{z}\bigr) \) into a do-free formula \( \mathcal{F}\bigl(P(\V)\bigr) \). Such a proof is said to \emph{hold in} a causal diagram~$\G$ if all the d-separation conditions required by each rule application are satisfied in~$\G$. In this case, the resulting expression \( \mathcal{F} \) is a valid causal estimand for the query and is referred to as an \emph{identification formula}.

Do-calculus is both sound and complete for causal diagrams~\citep{shpitser_identification_2006}: if a query is identifiable in a causal diagram~$\G$, then there exists a proof of do-calculus that holds in~$\G$. This proof yields a valid identification formula for the query in $\G$. Conversely, if the query is not identifiable in~$\G$, then no such a proof exists.

\section{Identifiability in a Collection of Graphs.}
\label{sec:collectionGraphs}

Specifying a full causal diagram demands detailed knowledge of all causal relationships among observed variables—an assumption rarely satisfied in practice.  To mitigate this challenge, many recent works have adopted \emph{causal abstractions}: simplified or coarse-grained models that capture only a subset of causal dependencies.

To our knowledge, each causal abstraction determines a collection $\C$ of causal diagrams on $\V$ that satisfy its specified constraints.  We study identifiability of a causal query within such a class: broadly, a query is \emph{identifiable} in $\C$ if the same estimand can be derived for every graph $G \in \C$.  In practice, however, this notion can be ambiguous.  The literature has proposed multiple identification strategies under causal abstractions, giving rise to several distinct definitions of identifiability in this setting.

\subsection{Epistemic Differences}

We now extend Pearl’s notion of identifiability from a single causal diagram to an entire class of diagrams.  Let $\C$ be a set of causal diagrams on the variable set~$\V$.  Intuitively, a causal query is identifiable in~$\C$ if it admits the same estimand for every graph in~$\C$.  We formalize this below.

\begin{definition}[Identifiability through Graphs (IG)]
\label{def:IG}
    Let $\C$ be a class of causal diagrams over~$\V$, and let $\X,\Y,\Z\subseteq\V$ be pairwise disjoint. The causal query 
    \(P\bigl(\mathbf{y} \mid \Do(\mathbf{x}),\mathbf{z}\bigr)\)
    is \emph{identifiable through graphs} in $\C$ if there exists a causal estimand valid for \(P\bigl(\mathbf{y} \mid \Do(\mathbf{x}),\mathbf{z}\bigr)\) in every  SCM $\M$ satisfying \(\G_{\M} \in \C\).
\end{definition}

Remark that it means that the causal query 
    \(P\bigl(\mathbf{y} \mid \Do(\mathbf{x}),\mathbf{z}\bigr)\)
    is \emph{identifiable through graphs} in $\C$ if there exists a causal estimand valid for \(P\bigl(\mathbf{y} \mid \Do(\mathbf{x}),\mathbf{z}\bigr)\) in all causal diagram $\G\in\C$ from Definition \ref{def:single}.

For instance, \citet{NEURIPS2022_17a9ab41} provide a complete characterization of identifiability through graphs in partial ancestral graphs.

The key distinction between identifiability in a single graph and identifiability through graphs lies in the set of SCMs under consideration.  In the single-graph setting, we restrict attention to those SCMs whose causal diagram is exactly $\G$.  By contrast, in the class-of-graphs setting, we consider the union of the SCM classes induced by every graph in~$\C$.  One may further restrict this union by incorporating additional domain knowledge.

As an illustrative example, suppose we know the true observational density \(P^\star(\V)\) (e.g.\ corresponding to an infinite sample size).  We then restrict our attention to those SCMs whose observational distribution satisfies 
\(
  P_{\M}(\V) = P^\star(\V).
\)
Under this constraint, we say a causal query is \emph{identifiable through Graphs knowing \(P^\star(\V)\)} (IGP) in the class \(\C\).

\begin{definition}[Identifiability through Graphs knowing $P^\star$ (IGP)]
\label{def:IGP}
Let $\C$ be a class of causal diagrams over~$\V$, and let $\X,\Y,\Z\subseteq\V$ be pairwise disjoint. Let $P^\star(\V)$ denote the true data density. The causal query 
\(P\bigl(\mathbf{y} \mid \Do(\mathbf{x}),\mathbf{z}\bigr)\)
is \emph{identifiable through graphs knowing $P^\star$} in $\C$ if there exists a causal estimand valid for \(P\bigl(\mathbf{y} \mid \Do(\mathbf{x}),\mathbf{z}\bigr)\) in every  SCM $\M$ satisfying \(\G_{\M} \in \C\) and $P_M(\V) = P^\star(\V)$.
\end{definition}

Despite their theoretical appeal, both definitions encounter serious practical limitations. Identifiability through graphs knowing $P^\star(\V)$ (IGP) requires knowing the true observational density—effectively demanding an infinite sample. IG can, in principle, be checked by enumerating every \(G\in\C\) and verifying that all derived estimands coincide; however, \(\C\) may be extremely large or even infinite, rendering this procedure infeasible.  To overcome these obstacles, we propose to distinguish between different notions of identifiability based on the proof techniques used to establish IG in \(\C\).

\subsection{Methodological Differences}

IG represents the strongest notion of graphical identifiability one can hope to achieve without making structural assumption or additional assumptions about the data distribution. In general, IG is a highly challenging problem. Hence, simpler notions of identifiability are proposed here.

To establish identifiability of a causal query in a collection \(\mathcal{C}\), one possible approach is to apply do‐calculus uniformly across all graphs in \(\mathcal{C}\). Specifically, this entails finding a single proof that holds in all causal diagram of the collection.

\begin{definition}[Identifiability by Common Do‐Calculus (ICD)]
\label{def:ICD}
    Let \(\mathcal{C}\) be a class of causal diagrams over the variable set \(\mathcal{V}\), and let \(\mathbf{X}, \mathbf{Y}, \mathbf{Z} \subseteq \mathcal{V}\) be pairwise disjoint. The causal query \(P\bigl(\mathbf{y}\mid \Do(\mathbf{x}),\,\mathbf{z}\bigr)\) is \emph{identifiable by common do‐calculus} in \(\mathcal{C}\) if there exists a proof of do‐calculus that holds in every \(\G \in \mathcal{C}\) and transforms the query into an expression involving only observational distributions.
\end{definition}

Moreover, we restate the notion of an \emph{atomically complete calculus}, introduced in \citet{NEURIPS2022_17a9ab41}, for a class \(\C\).\footnote{An example of an atomically complete calculus that turns out to be complete is given in \citet{anand_causal_2023}.} Such a calculus consists of three atomic rules $(A1, A2, A3)$, which are counterparts of the three do-calculus rules $(R1, R2, R3)$, and for each \(i\):
\begin{enumerate}
  \item \textbf{(Soundness)} If the atomic rule \(Ai\) applies to \(\C\), then the corresponding do-calculus rule \(Ri\) holds in every graph in \(\C\).
  \item \textbf{(Completeness)} If the atomic rule \(Ai\) does \emph{not} apply to \(\C\), then there exists at least one graph \(G \in \C\) in which the corresponding do-calculus rule \(Ri\) does not hold.
\end{enumerate}

Consequently, any derivation constructed in an atomically complete calculus immediately yields a common do‐calculus proof, and hence establishes ICD for the underlying class of diagrams. An example of a class where ICD has been achieved via this approach is that of Markov equivalence class of PAGs \citep{NEURIPS2022_17a9ab41}.

The main distinction between identifiability through graphs and identifiability by common do-calculus lies in the permissible proof techniques. In the latter, one must provide a single proof of do‐calculus that holds in all compatible graphs. In contrast, identifiability through graphs allows one to apply any graphical identification procedure on each compatible graph individually, so long as every such procedure yields the same, valid causal estimand.

Sometimes, it is also possible to identify an effect on an isolated graph using a \emph{graphical criterion} i.e. a set of graphical independence conditions that the diagram must satisfy in order for the criterion to apply. For instance, \cite{pearl_causality_2009} mentions the backdoor criterion and the frontdoor criterion. This leads to concepts such as \textit{identifiability by common backdoor (ICB)} or \textit{identifiability by common frontdoor (ICF)}, where one verifies that the graphical criterion, e.g., the backdoor criterion, applies to all graphs in $\mathcal{C}$. This defines a multitude of different graphical identifiability notions. The union of all them gives Definition \ref{def:ICGC}.

\begin{definition}(Identifiability by Common Graphical Criterion (ICGC))
\label{def:ICGC}
    Let \(\C\) be a class of causal diagrams over the variable set \(\V\), and let \(\X,\Y,\Z\subseteq\V\) be pairwise disjoint.  The causal query \(P\bigl(\mathbf{y}\mid \Do(\mathbf{x}),\,\mathbf{z}\bigr)\) is \emph{identifiable by common graphical criterion} in \(\C\) if there exists a graphical criterion satisfied in all graphs in $\mathcal{C}$.
\end{definition}

Example~\ref{ex:ICB} illustrates a class in which a query is identifiable by common backdoor. 

\begin{example}
\label{ex:ICB}
Let us consider the following class of causal diagram $\C$ as follows:
\[\C = 
\left\{ 
\begin{tikzpicture}[ ->, >=stealth, circle, baseline=(current bounding box.center), scale=.8, transform shape]
    % Nodes
    \node (X) at (0, 0) {X};
    \node (Y) at (1, 0) {Y};
    \node (Z) at (-1, 0) {Z};
    \node (U) at (-1.707, -0.707) {U};
    \node (V) at (-1.707, 0.707) {V};

    % Edges
    \draw (X) -- (Y);
    \draw (Z) -- (X);
    \draw (U) -- (Z);
    \draw (V) -- (Z);
    \draw (U) -- (V);
    \path (V) edge[bend left=25] (Y);
\end{tikzpicture}
,
\begin{tikzpicture}[ ->, >=stealth, circle, baseline=(current bounding box.center), scale=.8, transform shape]
    % Nodes
    \node (X) at (0, 0) {X};
    \node (Y) at (-1, 0) {Y};
    \node (Z) at (1, 0) {Z};
    \node (U) at (1.707, -0.707) {U};
    \node (V) at (1.707, 0.707) {V};

    % Edges
    \draw (X) -- (Y);
    \draw (Z) -- (X);
    \draw (U) -- (Z);
    \draw (V) -- (Z);
    \draw (V) -- (U);
    \path (U) edge[bend left=25] (Y);
\end{tikzpicture}
\right\}
\]

In both graphs of $\C$, $\{Z\}$ is a valid backdoor set. As a result, any causal query $P\left(y \mid \Do(x)\right)$ is identified by the formula $P\left(y \mid \Do(x)\right) = \sum_z P\left(y \mid x, z\right) P(z)$.

\end{example}

Identifiability by common backdoor in summary causal graphs has been studied in \citet{10.5555/3702676.3702683}, and was later extended into a complete criterion for identifiability by common adjustment by \citet{multivariateICA}. In a different framework, \citet{perkovic_complete_2018} established a sound and complete criterion for identifiability by common adjustment in Markov equivalence classes of ancestral graphs.

\section{Properties of Identifiability notions in collections of graphs}
\label{sec:prop} 

\subsection{A Property Valid for All Graphical Notions}

All notions of identifiability relying solely on the collection of graphs may be difficult to apply in practice because the collection \(\mathcal{C}\) can be very large. However, it is possible to reduce the size of \(\mathcal{C}\) without affecting the identifiability of the collection. Since edges in a graph represent dependencies, removing an edge preserves the graphical independences present in the original graph. This intuition is formalized in Lemma~\ref{lemma:Identifiabilite_et_inclusion}.

\begin{lemma}
    \label{lemma:Identifiabilite_et_inclusion}
    Let $\V$ denotes the set of observable variables. Let $\G_1$ and $\G_2$ be two causal diagrams on $\V$ and let \(\X,\Y,\Z\subseteq\V\) be pairwise disjoint. If $\G_2$ is a subgraph of $\G_1$, then the following propositions hold:
    \begin{enumerate}
        \item Any causal estimand valid in $\G_1$ for \(P\bigl(\mathbf{y}\mid \Do(\mathbf{x}),\,\mathbf{z}\bigr)\) is valid in $\G_2$ for \(P\bigl(\mathbf{y}\mid \Do(\mathbf{x}),\,\mathbf{z}\bigr)\).
        \item Any proof of do-calculus that holds in $\G_1$ also holds in $\G_2$.
        
        \item Any graphical criterion satisfied in $\G_1$ is satisfied in $\G_2$.
    \end{enumerate}
\end{lemma}

\begin{proof}
    Let us prove the different points:
    \begin{enumerate}
        \item  Let $\F_1$ be a causal estimand valid in $\G_1$ for \(P\bigl(\mathbf{y}\mid \Do(\mathbf{x}),\,\mathbf{z}\bigr)\). Let $\M_2$ be an SCM inducing $\G_2$. By adding missing arguments in the structural functions of $\M_2$, we can construct an SCM $\M_1$ inducing $\G_1$ such that $P_{\M_1}(\V) = P_{\M_2}(\V)$ and $P_{\M_1} \bigl(\mathbf{y} \mid \Do(\mathbf{x}), \mathbf{z}\bigr) = P_{\M_2}\bigl(\mathbf{y} \mid \Do(\mathbf{x}), \mathbf{z}\bigr)$:
        
        Indeed, for all $V_i \in \V$, we know that $\Pa(V_i,\G_2) \subseteq \Pa(V_i,\G_1)$. For any variable $A$, let $\mathcal{X}_A$ denote its valuation space.
        We construct $\M_1$ by keeping the same exogenous variables $\U$, endogenous variables $\V$, and exogenous distribution $P(\U)$ as in $\M_2$, and by defining new structural functions $f_i^{(1)} \in \F_{\M_1}$ for each $V_i \in \V$ as:
        \[
        f_i^{(1)} :
        \begin{cases}
        \mathcal{X}_{\Pa(V_i,\G_1)} \times \mathcal{X}_{\U_i} \to \mathcal{X}_{V_i} \\
        (x_{\Pa(V_i,\G_1)}, u_i) \mapsto f_i^{(2)}(x_{\Pa(V_i,\G_2)}, u_i)
        \end{cases}
        \]
        where $f_i^{(2)} \in \F_{\M_2}$ is the corresponding structural function in $\M_2$. That is, $f_i^{(1)}$ coincides with $f_i^{(2)}$ and ignores the arguments in $\Pa(V_i,\G_1) \setminus \Pa(V_i,\G_2)$. As a result, $\M_1$ and $\M_2$ induce the same observational and interventional distributions, while $\M_1$ respects the structural dependencies encoded in $\G_1$.

        Hence, $\F_1(P_{\M_2}(\V)) = \F_1(P_{\M_1}(\V)) =  P_{\M_1}\bigl(\mathbf{y} \mid \Do(\mathbf{x}), \mathbf{z}\bigr) = P_{\M_2} \bigl(\mathbf{y} \mid \Do(\mathbf{x}), \mathbf{z}\bigr)$. Therefore, $\F_1$ is valid in $\G_2$.
        
        \item Since $\G_2$ is a subgraph of $\G_1$, every graphical independence that holds in $\G_1$ also holds in $\G_2$. Therefore, the proof of do-calculus also holds in $\G_2$.
        
        \item Since $\G_2$ is a subgraph of $\G_1$, every graphical independence that holds in $\G_1$ also holds in $\G_2$. Therefore, any graphical criterion satisfied in $\G_1$ is satisfied in $\G_2$.
    \end{enumerate}
\end{proof}

Lemma~\ref{lemma:Identifiabilite_et_inclusion} implies that, to determine identifiability through graphs in the collection $\mathcal{C}$, it is sufficient to consider only the maximal graphs in $\mathcal{C}$ (with respect to subgraph inclusion). This is formalized in Theorem~\ref{th:identifiabilite_et_elements_maximaux}.

\begin{theorem}
\label{th:identifiabilite_et_elements_maximaux}
Let $\mathcal{C}$ be a class of causal diagrams over a set of variables~$\mathcal{V}$, and let $\mathbf{X}, \mathbf{Y}, \mathbf{Z} \subseteq \mathcal{V}$ be pairwise disjoint. Let \(\C^{\max}\) denote the subcollection of maximal elements of \(\C\) under graph inclusion. Consider any graphical notion of identifiability as described above (i.e. IG, ICD, ICGC, ICB, ICF). Then the following propositions are equivalent:
\begin{enumerate}
    \item \(P\bigl(\mathbf{y}\mid \Do(\mathbf{x}),\,\mathbf{z}\bigr)\) is identifiable in $\mathcal{C}$.
    \item  \(P\bigl(\mathbf{y}\mid \Do(\mathbf{x}),\,\mathbf{z}\bigr)\) is identifiable in \(\C^{\max}\).
\end{enumerate}
\end{theorem}

\begin{proof}
To treat all identifiability notions uniformly, we define an \emph{identification object} as either a causal estimand, a do‐calculus proof, or a graphical criterion, and say it \emph{applies} to \(\G\) if it is valid in \(\G\), holds in \(\G\), or is satisfied in \(\G\), respectively. Let us prove the two implications:

%\begin{itemize}
     \textbf{(1)} $ \Rightarrow$ \textbf{(2)}:  
    Suppose \(P\bigl(\mathbf{y} \mid \Do(\mathbf{x}),\, \mathbf{z}\bigr)\) is identifiable in $\mathcal{C}$. \(\C^{\max}\) is a subcollection of $\C$. Hence, the query is identifiable in \(\C^{\max}\).\\

     \textbf{(2)} $\Rightarrow$ \textbf{(1)}:
    Suppose now that \(P(\mathbf{y}\mid\Do(\mathbf{x}),\mathbf{z})\) is identifiable in \(\C^{\max}\). By definition, there exists an identification object \(\theta^{\max}\) applying to every \(\G^{\max}\in\C^{\max}\). Let $\G$ be any causal diagram in $\C$. Then, by definition of maximality, there exists $\G^{\max} \in \C^{\max}$ such that $\G$ is a subgraph of $\G^{\max}$. By Lemma~\ref{lemma:Identifiabilite_et_inclusion}, \(\theta^{\max}\) also applies to \(\G\). Then the query is identifiable in \(\C\). 
%\end{itemize}
\end{proof}

Theorem~\ref{th:identifiabilite_et_elements_maximaux} shows that identifying a causal effect becomes computationally easier when the subcollection of maximal elements under graph inclusion is small. As formalized in Corollary~\ref{cor:single_max}, if this subcollection reduces to a singleton, then identifiability in the entire class $\mathcal{C}$ reduces to identifiability in a single graph. 

\begin{corollary}
\label{cor:single_max}
    If $\mathcal{C}$ contains a greatest element under graphical inclusion, then IG is equivalent to ICD.
\end{corollary}

This situation arises, for example, in Cluster DAGs \citep{anand_causal_2023}, where the partition admissibility assumption ensures that there is a unique maximal element under inclusion. Designing causal abstractions that induce few maximal elements therefore appears to be a good practice when aiming for tractable identifiability conditions.

\begin{figure}[t]
    \centering
    \begin{tikzpicture}[  
        implies/.style={-Implies, double, double distance=2pt, thick},
        equiv/.style={Implies-Implies, double, double distance=2pt, thick},
        notimplies/.style={
            -Implies, double, double distance=.8pt, thick,
            postaction={
              decorate,
              decoration={
                markings,
                mark=at position 0.5 with {
                  \draw[-,line width=1pt, red] (-5pt,-5pt) -- (5pt,5pt);
                  \draw[-,line width=1pt, red] (-5pt,5pt) -- (5pt,-5pt);
                }
              }
            }
          },
        implies?/.style={
          -Implies, double, double distance=.8pt, thick,
          postaction={
            decorate,
            decoration={
              markings,
              mark=at position 0.5 with {
                \node[blue, left,font=\bfseries\huge] {?};
              }
            }
          }
        }]
    % Node
    \node (IGP)               {IGP};
    \node (IG)  [above=of IGP] {IG};
    \node (ICD) [above=of IG] {ICD};
    \node (ICGC) [above=of ICD] {ICGC};
    \node (ICF) [above right =of ICGC] {ICF};
    \node (ICB) [above left=of ICGC] {ICB};
    \node (dots) [above =of ICGC] {$\cdots$};
    
    % Arrows (implications)
    \draw[implies] (IG) -- (IGP);
    \draw[implies] (ICD) -- (IG);
    \draw[equiv] (ICD) -- (ICGC);
    \draw[implies] (ICB) -- (ICGC);
    \draw[implies] (ICF) -- (ICGC);
    \path (IGP) edge[notimplies, bend right =60, looseness = 1.2] (IG);
    \path (IG) edge[implies?, bend left =60, looseness = 1.2] (ICD);
    \path (ICD) edge[notimplies, bend left =20] (ICB);
    \path (ICD) edge[notimplies, bend right =20] (ICF);
    
    \end{tikzpicture}
    \caption{Relations between the notions of identifiability. An arrow indicates that the implication holds in every case. A red cross on an arrow indicates that the implication fails in at least one case. A blue question mark on an arrow indicates that the validity of the implication is still open.}
    \label{fig:liens_notions_id}
\end{figure}
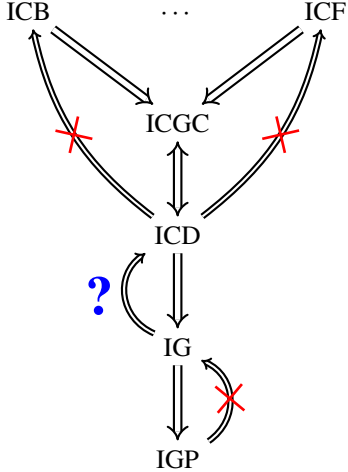

\subsection{Links Between These Concepts}

This section explores the fundamental relationships between the different notions of identifiability introduced earlier: Theorem~\ref{th:liens_notions_id} details the different implications and Figure~\ref{fig:liens_notions_id} illustrates all the logical implications that relate the various notions of identifiability discussed in this work.

\begin{theorem}
\label{th:liens_notions_id}
    Let $\C$ be a class of causal diagrams over~$\V$, and let $\X,\Y,\Z\subseteq\V$ be pairwise disjoint. We have the following properties:
    \begin{itemize}
        \item If \(P\bigl(\mathbf{y}\mid \Do(\mathbf{x}),\,\mathbf{z}\bigr)\) is identifiable by a common specific graphical criterion in $\C$ (e.g. ICB or ICF), then it is ICGC in $\C$.
        \item \textbf{ICGC $\Leftrightarrow$ ICD:} \(P\bigl(\mathbf{y}\mid \Do(\mathbf{x}),\,\mathbf{z}\bigr)\) is ICGC in $\C$ if and only if it is ICD in $\C$.
        \item \textbf{ICD $\Rightarrow$ IG:} if \(P\bigl(\mathbf{y}\mid \Do(\mathbf{x}),\,\mathbf{z}\bigr)\) is ICD in $\C$, then it is IG in $\C$.
        \item \textbf{IG $\Rightarrow$ IGP:} if \(P\bigl(\mathbf{y}\mid \Do(\mathbf{x}),\,\mathbf{z}\bigr)\) is IG in $\C$, then it is IGP in $\C$.
    \end{itemize}
\end{theorem}

\begin{proof}
{Let us prove the implications:}
\begin{itemize}
    \item Identifibility by a common specific (e.g backdoor or frontdoor) criterion is a special case of ICGC.
    
    \item \textbf{ICGC \(\Rightarrow\) ICD:}  
    any graphical criterion on a single graph can be established by a proof of do‐calculus.  Therefore, if a query is identifiable by a common graphical criterion in \(\C\), it is also identifiable by common do‐calculus in \(\C\).

    \item \textbf{ICD \(\Rightarrow\) ICGC:}  
    if a query is identifiable by common do‐calculus in \(\C\), then there exists a single do‐calculus proof that holds in every \(\G\in\C\).  Since each rule in this proof depends solely on graphical independence conditions, the conjunction of all these conditions defines a common graphical criterion satisfied by every \(\G\in\C\).

    \item \textbf{ICD $\Rightarrow$ IG:} the causal query is identifiable by common do-calculus. Therefore, there exists a proof of do-calculus that transforms the causal query \(P\bigl(\mathbf{y}\mid \Do(\mathbf{x}),\,\mathbf{z}\bigr)\) into a do-free formula $\mathcal{F}(P\left(\V\right))$, where $\F$ is a valid identification formula for every $\G\in \C$. Therefore, \(P\bigl(\mathbf{y}\mid \Do(\mathbf{x}),\,\mathbf{z}\bigr)\) is identifiable through graphs in $\C$.
    
    \item \textbf{IG $\Rightarrow$ IGP:} let $\mathcal{C}$ be a collection of graphs in which the causal query \(P\bigl(\mathbf{y}\mid \Do(\mathbf{x}),\,\mathbf{z}\bigr)\) is identifiable through graphs. By definition, the underlying class of SCMs in Definition \ref{def:IGP} is strictly smaller. Thus, IGP holds in $\mathcal{C}$ for this query.
\end{itemize}
\end{proof}

In a single graph, neither the backdoor criterion nor the frontdoor criterion is complete for identification. Consequently, ICB and ICF cannot be complete for ICD. Moreover, no existing graphical criterion is complete in a single graph, when consider causal diagrams (ADMGs). This suggests that no standard graphical criterion is likely to be complete for identifiability via common do-calculus. %Hence, it appears unlikely that any common graphical criterion could achieve completeness for identifiability by common do-calculus. 

To establish IGP, one can use knowledge on the true density of data. As a result, graphical knowledge may not be sufficient to establish IGP. That's why IG is not complete for IGP as shown in Example~\ref{ex:IGP_not_implies_IG}.

\begin{example}
\label{ex:IGP_not_implies_IG}
Let us consider the collection of causal diagrams $\C \coloneqq \{ X \rightarrow Y, X \leftarrow Y \}$ and  let us consider the causal query $P(Y = 1 \mid \Do(X = 1))$.

$P(Y = 1 \mid \Do(X = 1))$ \textbf{is not identifiable through graphs in $\C$.} Indeed, let us consider the SCM $\M_1$ defined by $X \sim \mathcal{B}\left(\frac{1}{2}\right)$ and $Y = X \otimes U$ where $U \sim \mathcal{B}\left(\frac{1}{10}\right)$ and $\otimes$ denotes the XOR operator. Moreover, let us consider the SCM $\M_2$ defined by $Y \sim \mathcal{B}\left(\frac{1}{2}\right)$ and $X = Y \otimes U$ where $U \sim \mathcal{B}\left(\frac{1}{10}\right)$. Then, one can verify that:
\[
P_{\M_1}(x, y) = \frac{9}{20} \cdot \mathbf{1}\{y = x\} + \frac{1}{20} \cdot \mathbf{1}\{y \ne x\} = P_{\M_2}(x, y)
\]
However, $P_{\M_1}(Y = 1 \mid \Do(X = 1)) = \frac{9}{10}$ and  $P_{\M_2}(Y = 1 \mid \Do(X = 1)) = \frac{1}{2}$. 
Therefore, $P(Y = 1 \mid \Do(X = 1))$ is not identifiable through graphs in $\C$.

$P(Y = 1 \mid \Do(X = 1))$ \textbf{may be identifiable through graphs knowing $P^\star(X, Y)$ in $\C$.} Indeed, let us consider the case where $X$ and $Y$ are independent in the true data distribution $P^\star(X,Y)$. Let $\M_1$ be an SCM inducing $X \rightarrow Y$ such that $P_{\M_1}(X,Y) = P^\star(X,Y)$. By the do-calculus, we know that $P_{\M_1}(Y = 1 \mid \Do(X = 1)) = P_{\M_1}(Y = 1 \mid X = 1) = P^\star(Y=1)$. Similarly, let $\M_2$ be an SCM inducing $X \leftarrow Y$ such that $P_{\M_2}(X,Y) = P^\star(X,Y)$. By the do-calculus, we know that $P_{\M_2}(Y = 1 | do(X=1)) = P^\star(Y=1)$. Therefore the causal estimand $\F \coloneqq P(X,Y) \mapsto P(Y = 1)$ is a valid causal estimand in all SCM $\M$ satisfying \(\G_{\M} \in \C\), $P_M(X,Y) = P^\star(X,Y)$, and  $P(Y = 1 \mid \Do(X = 1))$ is identifiable  through graphs knowing $P^\star(X, Y)$ in $\C$.
\end{example}

One implication, specifically the relationship between IG and ICD, remains unresolved. This open question is the subject of a conjecture presented in Section~\ref{section:conjecture}. 

The hierarchy established in Theorem~\ref{th:liens_notions_id} allows for a classification of the identifiability notions by increasing complexity. The more general a notion is, the more difficult it becomes to fully characterize it in a given collection of graphs.

\section{IG vs ICD conjecture}

In this section we focus on two notions of identifiability in collections of causal diagrams: identifiability through graphs and identifiability by common do-calculus.  Despite extensive study, the precise relationship between these notions remains unresolved.  In particular, it is not known whether there exists a family of causal diagrams \(\mathcal{C}\) and a causal query \(Q\) such that
\( Q \)  is IG in $\C$ but is not ICD in \(\C\).
We conjecture that such a class \(\mathcal{C}\) indeed exists, even though no concrete example is yet available.  Proving this conjecture would establish a strict separation between graphical identifiability and common do-calculus identifiability in causal abstractions.  Conversely, if no such class exists, it would follow that any calculus which is atomically complete on \(\mathcal{C}\) is in fact fully complete on \(\mathcal{C}\), thereby unifying these two notions of identifiability.

\paragraph{How to prove the conjecture.} 
To prove the conjecture, one may exhibit two causal diagrams \(\mathcal{G}_1\) and \(\mathcal{G}_2\), two do-calculus proofs \(\mathcal{P}_1\) and \(\mathcal{P}_2\), and a causal query \(P\bigl(\mathbf{y}\mid \Do(\mathbf{x}),\,\mathbf{z}\bigr)\) satisfying the following properties:
\begin{itemize}
  \item \(\mathcal{P}_1\) and \(\mathcal{P}_2\) are respectively valid in \(\mathcal{G}_1\) and \(\mathcal{G}_2\).
  \item \(\mathcal{P}_1\) and \(\mathcal{P}_2\) yield the same identification formula for \(P\bigl(\mathbf{y}\mid \Do(\mathbf{x}),\,\mathbf{z}\bigr)\).
  \item There is \emph{no} single do-calculus proof that is valid in both \(\mathcal{G}_1\) and \(\mathcal{G}_2\).
\end{itemize}

\paragraph{How to disprove the conjecture.} 
To disprove the conjecture, one must show that for \emph{every} class of causal diagrams \(\C\) and \emph{every} query \(P\bigl(\mathbf{y}\mid \Do(\mathbf{x}),\,\mathbf{z}\bigr)\), if the query is IG in \(\C\), then it is also ICD in \(\C\).  Equivalently, one must provide a uniform procedure that takes as input any class \(\C\) and a query IG in \(\C\), and outputs a do-calculus proof that holds in all \(\G\in\C\). 

\label{section:conjecture}
\section{Conclusion}
\label{sec:conclusion}

In this paper, we have introduced a unified framework for comparing the relative difficulty of various identifiability notions in causal abstractions, modelled as collections of causal diagrams. By arranging these notions into a hierarchy, we make explicit which criteria are strictly stronger than others. This organization allows one, for any given abstraction, to determine whether the strongest possible identifiability notion has been met, or if there remain stronger criteria yet to be achieved.

Looking ahead, it is essential to rigorously classify the algorithmic complexity associated with each identifiability notion, and to develop efficient procedures for verifying them. A pivotal open challenge is the IG versus ICD conjecture: determining whether there exists a class of causal diagrams in which a given causal query is ICD but not IG.

\bibliography{References}

\begin{thebibliography}{18}
\providecommand{\natexlab}[1]{#1}
\providecommand{\url}[1]{\texttt{#1}}
\expandafter\ifx\csname urlstyle\endcsname\relax
  \providecommand{\doi}[1]{doi: #1}\else
  \providecommand{\doi}{doi: \begingroup \urlstyle{rm}\Url}\fi

\bibitem[Anand et~al.(2023)Anand, Ribeiro, Tian, and Bareinboim]{anand_causal_2023}
Tara~V. Anand, Adele~H. Ribeiro, Jin Tian, and Elias Bareinboim.
\newblock Causal {Effect} {Identification} in {Cluster} {DAGs}.
\newblock \emph{Proceedings of the AAAI Conference on Artificial Intelligence}, 37\penalty0 (10):\penalty0 12172--12179, June 2023.
\newblock ISSN 2374-3468.
\newblock \doi{10.1609/aaai.v37i10.26435}.
\newblock URL \url{https://ojs.aaai.org/index.php/AAAI/article/view/26435}.

\bibitem[Assaad et~al.(2022)Assaad, Devijver, and Gaussier]{assaad2022}
Charles~K. Assaad, Emilie Devijver, and Eric Gaussier.
\newblock Discovery of extended summary graphs in time series.
\newblock In James Cussens and Kun Zhang, editors, \emph{Proceedings of the Thirty-Eighth Conference on Uncertainty in Artificial Intelligence}, volume 180 of \emph{Proceedings of Machine Learning Research}, pages 96--106. PMLR, 01--05 Aug 2022.
\newblock URL \url{https://proceedings.mlr.press/v180/assaad22a.html}.

\bibitem[Assaad et~al.(2024)Assaad, Devijver, Gaussier, G\"{o}ssler, and Meynaoui]{10.5555/3702676.3702683}
Charles~K. Assaad, Emilie Devijver, Eric Gaussier, Gregor G\"{o}ssler, and Anouar Meynaoui.
\newblock Identifiability of total effects from abstractions of time series causal graphs.
\newblock In \emph{Proceedings of the Fortieth Conference on Uncertainty in Artificial Intelligence}. JMLR.org, 2024.

\bibitem[Beckers and Halpern(2018)]{DBLP:journals/corr/abs-1812-03789}
Sander Beckers and Joseph~Y. Halpern.
\newblock Abstracting causal models.
\newblock \emph{CoRR}, abs/1812.03789, 2018.
\newblock URL \url{http://arxiv.org/abs/1812.03789}.

\bibitem[Geiger et~al.(2021)Geiger, Lu, Icard, and Potts]{NEURIPS2021_4f5c422f}
Atticus Geiger, Hanson Lu, Thomas Icard, and Christopher Potts.
\newblock Causal abstractions of neural networks.
\newblock In M.~Ranzato, A.~Beygelzimer, Y.~Dauphin, P.S. Liang, and J.~Wortman Vaughan, editors, \emph{Advances in Neural Information Processing Systems}, volume~34, pages 9574--9586. Curran Associates, Inc., 2021.
\newblock URL \url{https://proceedings.neurips.cc/paper_files/paper/2021/file/4f5c422f4d49a5a807eda27434231040-Paper.pdf}.

\bibitem[Geiger et~al.(2025)Geiger, Ibeling, Zur, Chaudhary, Chauhan, Huang, Arora, Wu, Goodman, Potts, and Icard]{geiger2025causalabstractiontheoreticalfoundation}
Atticus Geiger, Duligur Ibeling, Amir Zur, Maheep Chaudhary, Sonakshi Chauhan, Jing Huang, Aryaman Arora, Zhengxuan Wu, Noah Goodman, Christopher Potts, and Thomas Icard.
\newblock Causal abstraction: A theoretical foundation for mechanistic interpretability, 2025.
\newblock URL \url{https://arxiv.org/abs/2301.04709}.

\bibitem[Jaber et~al.(2022)Jaber, Ribeiro, Zhang, and Bareinboim]{NEURIPS2022_17a9ab41}
Amin Jaber, Adele Ribeiro, Jiji Zhang, and Elias Bareinboim.
\newblock Causal identification under markov equivalence: Calculus, algorithm, and completeness.
\newblock In S.~Koyejo, S.~Mohamed, A.~Agarwal, D.~Belgrave, K.~Cho, and A.~Oh, editors, \emph{Advances in Neural Information Processing Systems}, volume~35, pages 3679--3690. Curran Associates, Inc., 2022.
\newblock URL \url{https://proceedings.neurips.cc/paper_files/paper/2022/file/17a9ab4190289f0e1504bbb98d1d111a-Paper-Conference.pdf}.

\bibitem[Parviainen and Kaski(2017)]{PARVIAINEN2017110}
Pekka Parviainen and Samuel Kaski.
\newblock Learning structures of bayesian networks for variable groups.
\newblock \emph{International Journal of Approximate Reasoning}, 88:\penalty0 110--127, 2017.
\newblock ISSN 0888-613X.
\newblock \doi{https://doi.org/10.1016/j.ijar.2017.05.006}.
\newblock URL \url{https://www.sciencedirect.com/science/article/pii/S0888613X17303134}.

\bibitem[Pearl(2009)]{pearl_causality_2009}
Judea Pearl.
\newblock \emph{Causality}.
\newblock Cambridge University Press, Cambridge, 2 edition, 2009.
\newblock ISBN 978-0-521-89560-6.
\newblock \doi{10.1017/CBO9780511803161}.
\newblock URL \url{https://www.cambridge.org/core/books/causality/B0046844FAE10CBF274D4ACBDAEB5F5B}.

\bibitem[Perković et~al.(2018)Perković, Textor, Kalisch, and Maathuis]{perkovic_complete_2018}
Emilija Perković, Johannes Textor, Markus Kalisch, and Marloes~H. Maathuis.
\newblock Complete {Graphical} {Characterization} and {Construction} of {Adjustment} {Sets} in {Markov} {Equivalence} {Classes} of {Ancestral} {Graphs}.
\newblock \emph{Journal of Machine Learning Research}, 18:\penalty0 1--62, May 2018.
\newblock ISSN 1532-4435.
\newblock \doi{10.3929/ethz-b-000278021}.
\newblock URL \url{https://www.research-collection.ethz.ch/handle/20.500.11850/285660}.

\bibitem[Rischel and Weichwald(2021)]{pmlr-v161-rischel21a}
Eigil~F. Rischel and Sebastian Weichwald.
\newblock Compositional abstraction error and a category of causal models.
\newblock In Cassio de~Campos and Marloes~H. Maathuis, editors, \emph{Proceedings of the Thirty-Seventh Conference on Uncertainty in Artificial Intelligence}, volume 161 of \emph{Proceedings of Machine Learning Research}, pages 1013--1023. PMLR, 27--30 Jul 2021.
\newblock URL \url{https://proceedings.mlr.press/v161/rischel21a.html}.

\bibitem[Shpitser and Pearl(2006)]{shpitser_identification_2006}
Ilya Shpitser and Judea Pearl.
\newblock Identification of joint interventional distributions in recursive semi-{Markovian} causal models.
\newblock In \emph{proceedings of the 21st national conference on {Artificial} intelligence - {Volume} 2}, pages 1219--1226, Boston, Massachusetts, July 2006. AAAI Press.
\newblock ISBN 978-1-57735-281-5.

\bibitem[Wahl et~al.(2024)Wahl, Ninad, and Runge]{WahlNinadRunge+2024}
Jonas Wahl, Urmi Ninad, and Jakob Runge.
\newblock Foundations of causal discovery on groups of variables.
\newblock \emph{Journal of Causal Inference}, 12\penalty0 (1):\penalty0 20230041, 2024.
\newblock \doi{doi:10.1515/jci-2023-0041}.
\newblock URL \url{https://doi.org/10.1515/jci-2023-0041}.

\bibitem[Weinstein and Blei(2024)]{weinstein2024hierarchicalcausalmodels}
Eli~N. Weinstein and David~M. Blei.
\newblock Hierarchical causal models, 2024.
\newblock URL \url{https://arxiv.org/abs/2401.05330}.

\bibitem[Yvernes et~al.(2025)Yvernes, Devijver, and Gaussier]{multivariateICA}
Clément Yvernes, Emilie Devijver, and Eric Gaussier.
\newblock Complete characterization for adjustment in summary causal graphs of timeseries.
\newblock In \emph{UAI2025}, 2025.

\bibitem[Zennaro et~al.(2022)Zennaro, Turrini, and Damoulas]{zennaro2022computingoptimalabstractionstructural}
Fabio~Massimo Zennaro, Paolo Turrini, and Theodoros Damoulas.
\newblock Towards computing an optimal abstraction for structural causal models, 2022.
\newblock URL \url{https://arxiv.org/abs/2208.00894}.

\bibitem[Zennaro et~al.(2023)Zennaro, Turrini, and Damoulas]{10.24963/ijcai.2023/638}
Fabio~Massimo Zennaro, Paolo Turrini, and Theodoros Damoulas.
\newblock Quantifying consistency and information loss for causal abstraction learning.
\newblock In \emph{Proceedings of the Thirty-Second International Joint Conference on Artificial Intelligence}, IJCAI '23, 2023.
\newblock ISBN 978-1-956792-03-4.
\newblock \doi{10.24963/ijcai.2023/638}.
\newblock URL \url{https://doi.org/10.24963/ijcai.2023/638}.

\bibitem[Zennaro et~al.(2024)Zennaro, Bishop, Dyer, Felekis, Calinescu, Wooldridge, and Damoulas]{10.5555/3702676.3702869}
Fabio~Massimo Zennaro, Nicholas Bishop, Joel Dyer, Yorgos Felekis, Anisoara Calinescu, Michael Wooldridge, and Theodoros Damoulas.
\newblock Causally abstracted multi-armed bandits.
\newblock In \emph{Proceedings of the Fortieth Conference on Uncertainty in Artificial Intelligence}, UAI '24. JMLR.org, 2024.

\end{thebibliography}

\end{document}